\documentclass{article}

\usepackage{named}
\usepackage{amsmath,amssymb,latexsym}
\font\teneuf=eufm10
\font\seveneuf=eufm7
\font\fiveeuf=eufm5

\newfam\frakturfam
\textfont\frakturfam\teneuf
\scriptfont\frakturfam\seveneuf
\scriptscriptfont\frakturfam\fiveeuf

\newcommand{\lesdot}{\mathrel{\mathord{<}\!\!\raise 
0.8 pt\hbox{$\scriptstyle\circ$}}} 

\newcommand{\rest}{\upharpoonright}

\newcommand{\<}{{\langle}}
\renewcommand{\>}{{\rangle}}

\newcommand{\A}{{\mathcal A}}

\newcommand{\D}{{\mathbb D}}

\newcommand{\E}{{\mathbb E}}

\renewcommand{\L}{{\mathcal L}}  
\newcommand{\M}{{\mathcal M}}

\newtheorem{theorem}{Theorem}[section] 

\newtheorem{corollary}[theorem]{Corollary} 
 
\newtheorem{definition}[theorem]{Definition}

\newtheorem{lemma}[theorem]{Lemma}

\newcommand{\QED}{\hspace{0.2in}\vrule width 6pt height 6pt depth 0pt 
                  \vspace{0.1in}} 

\def\entailsi{\mathrel|\joinrel\sim}
\newcommand{\entails}{\vdash}

\def\vec{\overrightarrow}

\newcommand{\imp}{\Rightarrow}

\newcommand{\Proof}{{\sc Proof} \hspace{0.2in}}

\newtheorem{ea-stuff}{EA Comment/Question/Stuff}

\newsavebox{\subsubsecstr}
\newsavebox{\subsubsecstrb}

\newcommand{\explain}[2]{
  \settowidth{\mytextwidth}{$#1$}
  \settoheight{\mytextheight}{$#1$}
  \settoheight{\myvdotsheight}{$\vdots$}
  \addtolength{\myvdotsheight}{\mytextheight}
  \addtolength{\mytextheight}{\mytextheight}
  \addtolength{\mytextheight}{1pt}
  #1
  \hspace{-\mytextwidth}
  \makebox[0cm][l]{\raisebox{-\the\mytextheight}{\vdots}}
  \makebox[0cm][c]{\raisebox{-\the\myvdotsheight}{#2}}
  \hspace{\mytextwidth}
}
\newlength{\mytextwidth} \newlength{\mytextheight}
\newlength{\myvdotsheight}

  {\begin{list}{}{%
      \setlength{\labelwidth}{#1}%
      \setlength{\leftmargin}{\labelwidth}%
      \addtolength{\leftmargin}{\labelsep}}}%
  {\end{list}}

\def\pubnote#1{
\thispagestyle{myheadings}
\markboth{#1}{#1}
\setlength\headheight{10pt}   \setlength\headsep{10pt}
\def\thepage{}
}

\usepackage{kr02}

\title{Interpolation Theorems for Nonmonotonic Reasoning Systems}

\author{{\bf Eyal Amir}\\
Computer Science Division\\
University of California at Berkeley\\
Berkeley, CA 94720-1776, USA\\
eyal@cs.berkeley.edu}

\usepackage{boxedminipage}
\usepackage{graphicx}
\usepackage{psfrag}
\usepackage{varioref}
\usepackage{multicol}
\usepackage{times}

\begin{document}

\pubnote{Appears in \emph{9th Int'l Workshop on Non-Monotonic Reasoning (NMR '2002)}.}

\maketitle

\begin{abstract}
  \emph{Craig's interpolation theorem} \cite{Craig:1957} is an important
  theorem known for propositional logic and first-order logic.
  It says that if a logical formula $\beta$ logically follows from a
  formula $\alpha$, then there is a formula $\gamma$, including
  only symbols that appear in both $\alpha,\beta$, such that
  $\beta$ logically follows from $\gamma$ and $\gamma$ logically
  follows from $\alpha$. Such theorems are important and useful for
  understanding those logics in which they hold as well as for
  speeding up reasoning with theories in those logics.
%  $\alpha,\beta$ are formulae such that
%  $\alpha\imp\beta$ then there is a formula $\gamma$ that includes
%  only symbols that appear in both $\alpha$ and $\beta$, such that
%  $\alpha\imp\gamma$ and $\gamma\imp\beta$.
%
  In this paper we present interpolation theorems in this spirit for
  three nonmonotonic systems: circumscription, default logic and logic
  programs with the stable models semantics (a.k.a. answer set
  semantics). These results give us better understanding of those
  logics, especially in contrast to their nonmonotonic characteristics.
  They suggest that some \emph{monotonicity} principle holds despite
  the failure of classic monotonicity for these logics.
  Also, they sometimes allow us to use methods for the decomposition
  of reasoning
  %(e.g., \cite{AmirMcIlraith:kr2000})
  for these systems, possibly increasing their applicability and
  tractability. Finally, they allow us to build structured
  representations that use those logics.
\end{abstract}

%%%%%%%%%%%%%%%%%%%%%%%% Some initial settings %%%%%%%%%%%%%%%%%%%%%%%

%\pagestyle{fancy}
%\rhead{}
%%\lhead{Machinery for Elaborating Action}
%%\rhead{
%\cfoot{}

%%%%%%%%%%%%%%%%%%%%%%%%%%%%%%%%%%%%%%%%%%%%%%%%%%%%%%%%%%%%%%%%%%%

\section{Introduction}
\label{sec:introduction}

Craig's interpolation theorem \cite{Craig:1957} is an important
theorem known for propositional logic and first-order logic (FOL).
It says that if $\alpha,\beta$ are two logical formulae
and $\alpha\entails\beta$, then there is a formula
$\gamma\in\L(\alpha)\cap\L(\beta)$ such that $\alpha\entails\gamma$
and $\gamma\entails\beta$ (``$\entails$'' is the classical logical
deduction relation; $\L(\alpha)$ is the language of $\alpha$ (the set 
of formulae built with the nonlogical symbols of $\alpha$, $L(\alpha)$)).
Such interpolation theorems allow us to break inference into pieces
associated with sublanguages of the language of that theory
%\cite{AmirMcIlraith:kr2000}, 
\cite{McIlraithAmir:ijcai2001}, 
for those formal systems in which they hold.
%From it follows another useful property of
%the logical system called \emph{explicit definability}
%\cite{Beth:1953}.
In AI, these properties have been used to speed up inference
for constraint satisfaction systems (CSPs), propositional logic and FOL
(e.g., \cite{dechter-pearl88b,darwiche98,McIlraithAmir:ijcai2001,DechterRish:1994,Darwiche:aij1997,AmirMcIlraith:kr2000,Dechter:1999bucket}
and to build structured representations
\cite{darwiche98,Amir:oositcalc2000,Darwiche:aij1997}

In this paper we present interpolation theorems for three nonmonotonic
systems: \emph{circumscription} \cite{McC80}, \emph{default logic}
\cite{Reiter80} and \emph{logic programs} with the Answer Set semantics
\cite{GelLif91classical,GelLif88}.
%\cite{JACM::GelderRS1991} -- Well-Founded semantics.
In the nonmonotonic setup there are several interpolation theorems for
each system, with different conditions for applicability and different
form of interpolation. This stands in contrast to classical logic,
where Craig's interpolation theorem always holds. 
Our theorems allow us to use methods for the decomposition of reasoning 
(a-la \cite{AmirMcIlraith:kr2000,McIlraithAmir:ijcai2001})
under some circumstances for these systems, possibly increasing their
applicability and tractability for structured theories.
%
%In particular, some
%of our results are for \emph{cautious entailment} (regarding those
%formulae entailed in \emph{all} models), while others are for
%\emph{brave entailment} (those formulae entailed in \emph{at least
%  one} model).
%%extensions (default logic), stable models (logic
%%programs), or minimal models (circumscription).
%Also, in these nonmonotonic reasoning systems one can distinguish
%several different parts: the theory, the minimized predicates and the
%varied symbols (in circumscription); the facts and the defaults (in
%default logic); and the heads and bodies of the rules (in logic programs).
%Changes to these parts influence the kind of interpolation theorem
%that we get.
%
%Below are some of the main theorems shown in this paper.
% under some conditions
We list the main theorems that we show in this paper below,
omitting some of their conditions for simplicity.

For circumscription we show that, under some conditions,
$Circ[\alpha;P;Q]\models \beta$ iff there is some set of formulae
$\gamma\subseteq\L(\alpha)\cap\L(\beta)$ such that
$\alpha\models\gamma$ and $Circ[\gamma;P;Q]\models \beta$.  For
example, to answer $Circ[BlockW;block;L(BlockW)]\models on(A,B)$, we
can compute this formula $\gamma\in\L(\{block,on,A,B\})$ from $BlockW$
\emph{without applying circumscription}, and then solve
$Circ[\gamma;block;L(BlockW)]\models on(A,B)$ (where $\gamma$ may be
significantly smaller than $BlockW$).

For default logic, letting $\alpha\entailsi_D\beta$ mean that every
extension of $\<\alpha,D\>$ entails $\beta$ (\emph{cautious
  entailment}), we show that, under some conditions,
%  under some conditions
if $\alpha\entailsi_D\beta$, then there is a formula
% iff
$\gamma\in\L(\alpha\cup D)\cap\L(\beta)$ such that
$\alpha\entailsi_D\gamma$ and $\gamma\entailsi_D\beta$.
For logic programs we show that if $P_1,P_2$
% under some conditions,
are two logic programs
%with $head(P_2)\cap body(P_1)=\emptyset$
and $\varphi\in\L(P_2)$ such that $P_1\cup P_2\entailsi^b\varphi$,
then there is $\gamma\in\L(P_1)\cap\L(P_2)$ such that
$P_1\entailsi^b\gamma$ and $P_2\cup\gamma\entailsi^b\varphi$ (here
$\entailsi^b$ is the \emph{brave} entailment for logic programs).

% Some of these results are applicable for brave entailment as well.
% We leave treatment of \emph{brave
%  entailment} and theorems for a given extension or stable set for
%future work or other researchers.

This paper focuses on the form of the interpolation theorems that hold
for those nonmonotonic logics. We do not address the possible
application of these results to the problem of automated reasoning
with those logics. Nonetheless, we mention that direct application of
those results is possible along the lines already explored for
propositional logic and FOL in
\cite{AmirMcIlraith:kr2000,McIlraithAmir:ijcai2001}.

No interpolation theorems were shown for nonmonotonic reasoning
systems before this paper. Nonetheless, some of our theorems for
default logic and logic programs are close to the \emph{splitting
  theorems} of \cite{LifTur:1994,Turner:aaai1996}, which have already
been used to decompose reasoning for those logics.  The main
difference between our theorems and those splitting theorems is that
the latter change some of the defaults/rules involved to provide the
corresponding entailment. Also, they do not talk about an interpolant
$\gamma$, but rather discuss combining extensions.
%
%In circumscription, the reformulation of prioritized circumscription
%into several smaller circumscriptions (see \cite{Lif93}) is
%superficially reminiscent of ours in that both result in
%decompositions, but the the former are not interpolation theorems.
%
%he closest results to ours are possibly those pertaining to
%prioritized versions of these nonmonotonic systems.  For example,
%Prioritized circumscription, in which we minimize several predicates
%simultaneously, but prioritize the minimization process, e.g., giving
%precedence to minimizing $P_1$ over $P_2$, has an decomposed form that
%is equivalent to it (see e.g., \cite{Lif93}). These \emph{separation}
%theorems differ from ours in that our interpolation theorems depend on
%the query and also typically reduce the computation into two
%computations with theories in reduced languages.  Our theorems apply
%to prioritized cases as well as to non-prioritized ones. 
%
%A relationship between FOL and propositional logic interpolation
%theorems and nonmonotonic reasoning systems was first explored by
%\cite{JAR::Doyle1985}, without offering interpolation theorems for
%nonmonotonic reasoning systems.
%

Since its debut, the nonmonotonic reasoning line of work
has expanded and several textbooks now exist that give a fair view of
nonmonotonic reasoning and its uses (e.g.,
%\cite{Brewka91,Antoniou97,BreDixKon97,
\cite{HBLAI3}).
%,San94fe,Shanahan:1997book
%The motivations for nonmonotonic reasoning vary
%from formalizing Common Sense reasoning through Elaboration Tolerance
%and representation of uncertainty to Belief Revision. See, e.g., 
%\cite{Shanahan:1997book,McC98:cs98,San94fe}
%%Pearl-uncertainty,Antoniou97
%for further details in these directions.
The reader is referred to those books for background and further details.

%\emph{[Some proofs are shortened or omitted for lack of space. They appear in
%the final version.]}

%%%%%%%%%%%%%%%%%%%%%%%%%%%%%%%%%%%%%%%%%%%%%%%%%%%%%%%%%%%%%%%%%%%%%

\section{Logical Preliminaries}

In this paper, we use the notion of \emph{logical theory} for every
set of axioms in FOL or propositional logic, regardless of whether the
set of axioms is deductively closed or not.  We use $L(\A)$ to denote
the signature of $\A$, i.e., the set of non-logical symbols. $\L(\A)$
denotes the language of $\A$, i.e., the set of formulae built with
$L(\A)$.  $Cn(\A)$ is the set of logical consequences of $\A$ (i.e.,
those formulae that are valid consequences of $\A$ in FOL).
For a first-order structure, $M$, in $L$, we write $U(M)$ for the
universe of elements of $M$. For every symbol, $s$, in $L$, we write
$s^M$ for the interpretation of $s$ in $M$.

Finally, we note Craig's Interpolation Theorem.
\begin{theorem}[\cite{Craig:1957}]
  \label{thm:craig-interpol}
  Let $\alpha,\beta$ be sentences such that 
  $\alpha\entails\beta$. Then there is a formula $\gamma$ involving 
  only nonlogical symbols common to both $\alpha$ and $\beta$,
  such that $\alpha\entails \gamma$ and $\gamma\entails \beta$.
\end{theorem}

% Define Prime implicates????

%%%%%%%%%%%%%%%%%%%%%%%%%%%%%%%%%%%%%%%%%%%%%%%%%%%%%%%%%%%%%%%%%%%%%

\section{Circumscription}

%=====================================================================

%\section{Reasoning with circumscription}
\subsection{McCarthy's Circumscription: Overview}
\label{sec:circ-bg}

McCarthy's circumscription \cite{McC80,McC86} is a nonmonotonic
reasoning system in which inference from a set of axioms, $A$, is
performed by minimizing the extent of some predicate symbols
$\vec{P}$, while allowing some other nonlogical symbols, $\vec{Z}$ to
vary.

Formally, McCarthy's circumscription formula %\cite{McC86}
\begin{equation}
  \label{eq:circ-def}
  \begin{array}{l}
  Circ[A(P,Z);P;Z]=\\
  \ \ \ A(P,Z)\land \forall p,z\ (A(p,z)\imp \neg(p<P))
\end{array}
\end{equation}
says that in the theory $A$, with parameter relations and function
vectors (sequence of symbols) $P,Z$, $P$ is a minimal element such
that $A(P,Z)$ is still consistent, when we are allowed to vary $Z$ in
order to allow $P$ to become smaller.

Take for example the following simple theory:
\[\begin{array}{l}
  T\ =\ block(B_1)\land block(B_2)
\end{array}\]
Then, the circumscription of $block$ in $T$, varying nothing, is
\[Circ[T;block;]=T\land\forall p\ [T_{[block/p]}\imp \neg(p<block)].\]
Roughly, this means that $block$ is a minimal predicate satisfying $T$.
Computing circumscription is discussed in length in \cite{Lif93} and
others, and we do not expand on it here. Using known techniques we can
conclude
\[Circ[T;block;]\equiv \forall x\ (block(x)\Leftrightarrow (x=B_1\vee x=B_2))\]
This means that there are no other blocks in
the world other than those mentioned in the original theory $T$.

%=====================================================================

%------------------------------------------------------------------------

%\subsubsection{Semantics for circumscription}
%\label{sec:circ-semantics}

We give the preferential semantics for circumscription that was given
by \cite{Lif85,McC86,Eth86} in the following definition.
\begin{definition}[\cite{Lif85}]
  For any two models $M$ and $N$ of a theory $T$ we write $M\leq_{P,Z}
  N$ if the models $M,N$ differ only in how they interpret predicates
  from $P$ and $Z$ and if the extension of every predicate from $P$ in
  $M$ is a subset of its extension in $N$. We write $M<_{P,Z}N$ if for
  at least one predicate in $P$ the extension in $M$ is a strict
  subset of its extension in $N$.
\end{definition}
%
%The relation $\leq_{P,Z}$ is a pre-order and hence we can talk about
%minimal models $M$ with respect to $\leq_{P,Z}$ in the class of all
%models of $T$. More precisely, 
We say that a model $M$ of $T$ is
\emph{$\leq_{P,Z}$-minimal} if there is no model $N$ such that
$N<_{P,Z}M$.

\begin{theorem}[\cite{Lif85}: Circumscript. Semantics]
  \label{thm:circ-semantics}
  Let $T$ be a finite set of sentences.  A structure $M$ is a model of
  $Circ[T;P;Z]$ iff $M$ is a $\leq_{P,Z}$-minimal model of $T$.
\end{theorem}

This theorem allows us to extend the definition of circumscription to
set of infinite number of sentences. In those cases, $Circ[T;P;Z]$ is
defined as the set of sentences that hold in all the $\leq_{P,Z}$-minimal
models of $T$. Theorem \ref{thm:circ-semantics} implies that this
extended definition is equivalent to the syntactic characterization of
the original definition (equation (\ref{eq:circ-def})) if $T$ is a
finite set of sentences.  In the rest of this paper, we refer to this
extended definition of circumscription, if $T$ is an infinite set of
FOL sentences (we will note those cases when we encounter them).

%A similar theorem is given with respect to the semantics for
%Prioritized circumscription.
%\begin{theorem}[\cite{Lif85}: Prioritized-Circ. Semantics]
%  A structure $M$ is a model of the prioritized circumscription
%  $Circ[T;P_1>...>P_n;Z]$ iff $M$ is a model of $T$ such that for
%  every $i\leq n$, $M$ is $(P_i,\{P_{i+1},...,P_n,Z\})$-minimal model of $T$.
%\end{theorem}

%This last theorem is also a direct consequence of the previous theorem 
%and theorem \vref{thm:circ-prioritized-circ}.

Circumscription satisfied Left Logical Equivalence (LLE): 
$T\equiv T'$ implies that $Circ[T;P;Z]\equiv Circ[T';P;Z]$. It also
satisfies Right Weakening (RW):
$Circ[T;P;Z]\models\varphi$ and $\varphi\imp\psi$
implies that $Circ[T;P;Z]\models\psi$).

%\subsection{Semantics for circumscription}

%In \cite{Lif85}, circumscription was given semantic interpretation.

%\begin{definition}
%\label{def:leq-xi}
%  Let $\M_1,\M_2$ have the same universe $U$.
%  %and let $\xi\in U^k$, where
%  %$k$ is the arity of a predicate symbol $S_1$.
%  We say that $\M_1\leq^{P,Q}\M_2$ iff
%  \begin{enumerate}
%  \item $K^{\M_1}=K^{\M_2}$ for every function, predicate or constant
%    symbol, $K$, that is not in $P,Q$,
%  \item $K^{\M_1}(\alpha)\implies K^{\M_2}(\alpha)$, for every
%    predicate $K$ in $P$ and every $\xi\in U^k$ ($k$ is the arity of $K$).
%  \end{enumerate}
%\end{definition}

%\begin{proposition}[\cite{Lif85}]
%\label{prop:semantics-circ}
%Let $\M$ be a model of $A(S)$.
%\[\begin{array}{l}
%  \M\models C_{PW}[A;{S_1};S_1/V_1,...,S_n/V_n]\iff\\
%    \hspace{0.2in}\forall\M'\in\lbv A(S)\rbv\ \forall\xi\in |\M|^k\ \\
%    \hspace{1in}    \neg(\M'\leq^\xi\M\land \M\not\leq^\xi\M')
%\end{array}\]
%\end{proposition}

%Other words, $\M\models C_{PW}[A;{S_1};S_1/V_1,...,S_n/V_n]$ iff for each
%$\xi\in |\M|^k$, $\M$ is minimal relative to $\leq^\xi$.

%======================================================================

\subsection{Model Theory}

%The following reviews some classical results in \emph{model theory}
%(see \cite{Hodges:book1997} for a reference).

\begin{definition}
  Let $M,N$ be $L$-structures, for FOL signature $L$ and language
  $\L$.  We say that $N$ is an \emph{elementary extension} of $M$ (or
  $M$ is an \emph{elementary substructure} of $N$), written $M\preceq
  N$, if $U(M)\subseteq U(N)$ and for every $\varphi(\vec{x})\in\L$
  and vector of elements $\vec{a}$ of $U(M)$,
  $M\models\varphi(\vec{a})$ iff $N\models\varphi(\vec{a})$.
\end{definition}
% Hodges p.48

$f:M\rightarrow N$ is an \emph{elementary embedding} if $f$ is an
injective (one-to-one) homomorphism from $M$ to $N$ and for every
$\varphi(\vec{x})\in\L$ and vector $\vec{a}=\<a_1,...,a_n\>$ of
elements from $U(M)$ (i.e., $a_1,..,a_n\in U(M)$),
$M\models\varphi(\vec{a})$ iff $N\models \varphi(f(a_1),...,f_(a_n))$.

For FOL signatures $L\subseteq L^+$, and for $N$ an $L^+$-structure, we
say that $N\rest L$ is the \emph{reduct} of $N$ to $L$, the
$L$-structure with the same universe of elements as $N$, and the same
interpretation as $N$ for those symbols from $L^+$ that are in $L$
(there is no interpretation for symbols not in $L$).
% Hodges p.9
For A theory $T$ in a language of $L^+$, let $Cn^L(T)$ be the set of
all consequences of $T$ in the language of $L$.

The following theorem is a model-theoretic property that is analogous
to Craig's interpolation theorem (Theorem \ref{thm:craig-interpol}).
\begin{theorem}[See \cite{Hodges:book1997} p.148]
  \label{thm:lang-extension}
  Let $L,L^+$ be FOL signatures with $L\subseteq L^+$ and $T$ a theory 
  in the language of $L^+$. Let $M$ be an $L$-structure.
  Then, $M\models Cn^L(T)$ if and only if for some model $N$ of $T$,
  $M\preceq N\rest L$ ($M$ is an elementary substructure of the reduct
  of $N$ to $L$).
\end{theorem}

%\begin{theorem}[Heir-coheir amalgams]
%  \label{thm:heir-coheir}
%  Let $A,B,C$ be structures such that $A\preceq B$ and $A\preceq C$.
%  Then there exist an elementary extension $D$ of $B$ and an
%  elementary embedding $g:C\rightarrow D$ such that $g(C)\preceq D$.
%\end{theorem}
% Hodges p.137

%=========================================================================

\subsection{Interpolation in Circumscription}
\label{sec:interpolation-circ}

In this section we present two interpolation theorems for
circumscription.  Those theorems hold for both FOL and propositional
logic.  Roughly speaking, the first (Theorem \ref{thm:interpol-circ})
says that if $\alpha$ nonmonotonically entails $\beta$ (here this
means $Circ[\alpha;P;Q]\models\beta$), then there is
$\gamma\subseteq\L(\alpha)\cap\L(\beta\cup P)$ such that $\alpha$
classically entails $\gamma$ ($\alpha\models\gamma$) and $\gamma$
nonmonotonically entails $\beta$ ($Circ[\gamma;P;Q]\models\beta$).  In
the FOL case this $\gamma$ can be an infinite set of sentences, and we
use the extended definition of Circumscription for infinite sets of
axioms for this statement.

The second theorem (Theorem \ref{thm:interpol-circ2}) is similar to
the first, with two main differences. First, it requires that
$L(\alpha)\subseteq (P\cup Q)$. Second, it guarantees that $\gamma$ as
above (and some other restrictions) exists iff $\alpha$
nonmonotonically entails $\beta$. This is in contrast to the first
theorem that guarantees only that \emph{if} part.
The actual technical details are more fine than those rough
statements, so the reader should refer to the actual theorem
statements.

In addition to these two theorems, we present another theorem that
addresses the case of reasoning from the union of theories (Theorem
\ref{thm:circ-prime-implicate}). 
Before we state and prove those theorems, we prove several useful
lemmas.  

Our first lemma says that if we are given two theories $T_1,T_2$, and
we know the set of sentences that follow from $T_2$ in the
intersection of their languages, then every model of this set of
sentences together with $T_1$ can be extended to a model of $T_1\cup
T_2$.

\begin{lemma}
  \label{lem:interpolation-model}
  Let $T_1,T_2$ be two theories, with signatures in $L_1,L_2$,
  respectively. Let $\gamma$ be a set of sentences logically equivalent
  to $Cn^{L_1\cap L_2}(T_2)$.  For every $L_1$-structure, $\M$, that
  satisfies $T_1\cup\gamma$ there is a $(L_1\cup L_2)$-structure,
  $\widehat{\M}$, that is a model of $T_1\cup T_2$ such that
  $\M\preceq \widehat{\M}\rest L_1$.
\end{lemma}

\Proof
Let $\M$ be a $L_1$-structure that is a model of
$T_1\cup\gamma$. Then $\M\models\gamma$. Noticing that $\gamma$ is
logically equivalent to $Cn^{L_1\cap L_2}(T_2)$ (by definition of
$\gamma$), we get that $\gamma\models Cn^{L_1\cap L_2}(T_2)$.
Consequently, $\gamma\models Cn^{L_1}(T_2)$ because
$Cn^{L_1\cap L_2}(T_2)\equiv Cn^{L_1}(Cn^{L_1\cap L_2}(T_2))=Cn^{L_1}(T_2)$.

Now we use Theorem \ref{thm:lang-extension} with $L=L_1$,
$L^+=L_1\cup L_2$, $M=\M$ and $T=T_1\cup T_2$. We know that
$\M\models T_1\cup\gamma$. Thus, $M\models T_1\cup Cn^{L_1}(T_2)$.
To use Theorem \ref{thm:lang-extension} we need to show that
$M\models Cn^{L_1}(T_1\cup T_2)$. 
We use Craig's interpolation theorem (Theorem
\ref{thm:craig-interpol}) to show this is indeed the case.

First notice that 
$Cn^{L_1}(T_1\cup T_2)\supseteq Cn^{L_1}(T_1\cup \gamma)$ is true
because $T_2\models\gamma$. We show that 
$Cn^{L_1}(T_1\cup T_2)\subseteq Cn^{L_1}(T_1\cup \gamma)$.
Take $\varphi\in Cn^{L_1}(T_1\cup T_2)$. By definition,
$T_1\cup T_2\models\varphi$ and $\varphi\in\L(T_1)$. The deduction
theorem for FOL implies that $T_2\models T_1'\imp\varphi$, for some
finite subset $T_1'\subseteq T_1$.
Craig's interpolation theorem for FOL implies that there is
$\delta\in\L(T_2)\cap\L(T_1'\imp\varphi)=\L(T_2)\cap\L(T_1)$
such that $T_2\models\delta$ and $\delta\models T_1'\imp \varphi$.
Thus, $\delta\in Cn^{L_1\cap L_2}(T_2)\equiv\gamma$. Consequently,
$\gamma\models T_1'\imp \varphi$. Using the deduction theorem again we
get that $T_1'\cup\gamma\models\varphi$, implying that
$T_1\cup\gamma\models\varphi$. 

Thus, we showed that
$Cn^{L_1}(T_1\cup T_2)=Cn^{L_1}(T_1\cup \gamma)$.
From $M\models T_1\cup Cn^{L_1}(T_2)$ and $\gamma=Cn^{L_1}(T_2)$ 
we get that $M\models Cn^{L_1}(T_1\cup T_2)$. 

Finally, the conditions of Theorem \ref{thm:lang-extension} for
$L=L_1$, $L^+=L_1\cup L_2$, $M=\M$ and $T=T_1\cup T_2$ hold.
We conclude that there is a
$(L_1\cup L_2)$-structure, 
$\widehat{\M}$, that is a model of $T_1\cup T_2$ such that
$\M\preceq\widehat{\M}\rest L_1$.  
\QED

Our second lemma says that every $<_{P,Q}$-minimal model of $T$ that
is also a model of $T'$ is a $<_{P,Q}$-minimal model of $T\cup T'$.

\begin{lemma}
  \label{lem:circ-extending-theory}
  Let $T$ be a theory and $P,Q$ vectors of nonlogical symbols.
  If $\M\models Circ[T;P;Q]$ and $\M\models T\cup T'$, then
  $\M\models Circ[T\cup T';P;Q]$.
\end{lemma}

\Proof
Let $\M$ be a model of $T\cup T'$ such that $\M\models Circ[T;P;Q]$.
If there is $\M'<_{P,Q}\M$ such that $\M'\models T\cup T'$, then
$\M'\models T$ and $\M\not\models Circ[T;P;Q]$. Contradiction.
Thus, there is no such $\M'$ and $\M\models Circ[T\cup T';P;Q]$.
\QED

The following theorem is central to the rest of our results in this
section. It says that when we circumscribe $P,Q$ in $T_1\cup T_2$ we
can replace $T_2$ by its consequences in $\L(T_1)$, for some purposes
and under some assumptions.

\begin{theorem}
  \label{thm:circ-T-gamma}
  Let $T_1,T_2$ be two theories and $P,Q$ two vectors of symbols from
  $L(T_1)\cup L(T_2)$ such that $P\subseteq L(T_1)$. Let $\gamma$ a set
  of sentences logically equivalent to $Cn^{L(T_1)\cap L(T_2)}(T_2)$.
%  Let $\gamma$ be the set of prime implicates of $T_2$ in $L(T_1)\cap L(T_2)$.
  Then, for all $\varphi\in\L(T_1)$, if
  $Circ[T_1\cup T_2;P;Q]\models\varphi$, then
  $Circ[T_1\cup\gamma;P;Q]\models\varphi$.
\end{theorem}

\Proof
We show that for every model of $Circ[T_1\cup\gamma;P;Q]$
there is a model of $Circ[T_1\cup T_2;P;Q]$ whose reduct to 
$L(T_1)$ is an elementary extension of the reduct of the first model
to $L(T_1)$.

Let $\M$ be a $L(T_1\cup T_2)$-structure that
is a model of $Circ[T_1\cup\gamma;P;Q]$. Then,
$\M\models T_1\cup\gamma$. 
From Lemma \ref{lem:interpolation-model} we know that there is 
a $(L_1\cup L_2)$-structure, $\widehat{\M}$, that is a model of $T_2$ such
that $\M\rest L(T_1) \preceq \widehat{\M}\rest L(T_1)$.

Thus, $\widehat{\M}$ is a $\leq_{P,Q}$-minimal model of
$T_1\cup\gamma$. To see this, assume otherwise. Then, there is a
model $\M'$ for the signature $L(T_1\cup T_2)$ such that
$\M'<_{P,Q}\widehat{\M}$ and $\M'\models T_1\cup\gamma$. 
Take $\M''$ such that the interpretation of all the symbols
in $L(T_1)$ is exactly the same as that of $\M'$ and such that
the interpretation of all symbols in $L(T_2)\setminus L(T_1)$
is exactly the same as that of $\M$. Then,
$\M''\models T_1\cup\gamma$ because
$T_1\cup\gamma\subseteq \L(T_1)$. Also, $\M''<_{P,Q'}\M$,
for $Q'=Q\cap L(T_1)$ because $P\subseteq L(T_1)$ and
$\M,\widehat{\M}$ agree on the interpretation of symbols in $L(T_1)$
($\M\rest L(T_1)\preceq \widehat{\M}\rest L(T_1)$).
Thus, $\M''<_{P,Q}\M$, since $\M'',\M$ agree on all the interpretation
of all symbols in $L(T_2)\setminus L(T_1)$.
This contradicts $\M\models Circ[T_1\cup\gamma;P;Q]$, so
$\widehat{\M}$ is a $\leq_{P,Q}$-minimal model of $T_1\cup\gamma$.

Thus, $\widehat{\M}\models Circ[T_1\cup \gamma;P;Q]$, and
$\widehat{\M}\models T_1\cup T_2$.
From Lemma \ref{lem:circ-extending-theory} we get that
$\widehat{\M}\models Circ[T_1\cup T_2;P;Q]$.

Now, let $\varphi\in \L(T_1)$ such that 
$Circ[T_1\cup T_2;P;Q]\models\varphi$. Then every model of
$Circ[T_1\cup T_2;P;Q]$ satisfies $\varphi$. Let $\M$ be a model
of $Circ[T_1\cup\gamma;P;Q]$ in the language $\L(T_1\cup T_2)$.
Then there is $\widehat{\M}$ as above, i.e.,
$\widehat{\M}\models Circ[T_1\cup T_2;P;Q]$ and
$\M\rest L(T_1) \preceq \widehat{\M}\rest L(T_1)$.
Thus, $\widehat{\M}\models\varphi$. Since
$\M\rest L(T_1)\preceq\widehat{\M}\rest L(T_1)$,
$\M\models\varphi$. Thus every model of $Circ[T_1\cup\gamma;P;Q]$
is a model of $\varphi$.
\QED

\begin{theorem}[Interpolation for Circumscription 1]
  \label{thm:interpol-circ}
  Let $T$ be a theory, $P,Q$ vectors of symbols, and $\varphi$ a formula.
%  $P\subseteq L(T\cup\{\varphi\})$ such that $(P\cup Q)\supseteq L(T)$.
  If $Circ[T;P;Q]\models\varphi$, then there is
  $\gamma\subseteq\L(T)\cap\L(\varphi\cup P)$ such that
  \[T\models\gamma \mbox{ \ \ and \ \ } Circ[\gamma;P;Q]\models\varphi.\]
  Furthermore, this $\gamma$ can be logically equivalent to the
  consequences of $T$ in $L(T)\cap L(\varphi\cup P)$.
\end{theorem}

\Proof
We use Theorem \ref{thm:circ-T-gamma} to find this $\gamma$. For
$T,\varphi$ as in the statement of the theorem we define $T_1,T_2$ as
follows. We choose $T_1$ such that $\varphi\in\L(T_1)$ and
$P\subseteq L(T_1)$:
Let $T_1=\{\varphi\vee\neg\varphi\}\cup\tau_1$ for $\tau_1$ a
set of tautologies such that $L(\tau_1)=P$. 
We choose $T_2$ such that it includes $T$ and has a rich enough
vocabulary so that $P,Q\subseteq L(T_1)\cup L(T_2)$.
Let $T_2=T\cup\tau_2$, for $\tau_2$ a set of tautologies such that
$L(\tau_2)=Q\setminus L(T_1)$.  Let $L_1=L(T_1)$, $L_2=L(T_2)$.

%Assume that $P\subseteq L(\varphi)$.
Theorem \ref{thm:circ-T-gamma} guarantees that if $P\subseteq L_1$
then $\gamma$ from that theorem satisfies
$Circ[T_1\cup T_2;P;Q]\models\psi \imp Circ[T_1\cup\gamma;P;Q]\models\psi$
for every $\psi\in\L(T_1)$. This implies that for every 
$\psi\in\L(\{\varphi\}\cup\tau_1)$,
$Circ[T;P;Q]\models\psi \imp Circ[\gamma;P;Q]\models\psi$.
In particular, $Circ[\gamma;P;Q]\models\varphi$, and this $\gamma$
satisfies our current theorem.
\QED

This theorem does not hold if we require
$\gamma\subseteq\L(T)\cap\L(\varphi)$ instead of
$\gamma\subseteq\L(T)\cap\L(\varphi\cup P)$. For example, take
$\varphi=Q$, $T=\{\neg P\imp Q\}$, where $P,Q$ are propositional
symbols.  $Circ[T;P;Q]\models\varphi$. However, every logical
consequence of $T$ in $L(\varphi)$ is a tautology.
Thus, if the theorem was correct with our changed requirement,
$\gamma$ would be equivalent to $\emptyset$ and
$Circ[\gamma;P;Q]\not\models\varphi$.

\begin{theorem}
  \label{thm:circ-gamma-T}
  Let $T_1,T_2$ be two theories, $P,Q$ two vectors of symbols from
  $L(T_1)\cup L(T_2)$ such that $P\subseteq L(T_1)$ and
  $P\cup Q\supseteq L(T_2)$. Let $\gamma$ be a set of sentences
  logically equivalent to $Cn^{L(T_1)\cap L(T_2)}(T_2)$.
%Let $\gamma$ be the set of prime implicates of $T_2$ in $L(T_1)\cap L(T_2)$.
  Then, for all $\varphi\in\L(T_1)$,
  if $Circ[T_1\cup \gamma;P;Q]\models\varphi$, then
  $Circ[T_1\cup T_2;P;Q]\models\varphi$.
\end{theorem}

\Proof
We show that every model of $Circ[T_1\cup T_2;P;Q]$
is also a model of $Circ[T_1\cup\gamma;P;Q]$.
Let $\M$ be a $L(T_1\cup T_2)$-structure that is a model of
$Circ[T_1\cup T_2;P;Q]$. Then $\M\models T_1\cup T_2$, implying that
also $\M\models T_1\cup\gamma$.

Assume that there is $\M'<_{P,Q}\M$ such that
$\M'\models T_1\cup\gamma$. From Lemma \ref{lem:interpolation-model},
there is $\bar{\M'}$ such that $\bar{\M'}\models T_1\cup T_2$ and
$\M'\rest L(T_1) \preceq \bar{\M'}\rest L(T_1)$.
Since $\M',\bar{\M'}$ agree on all the symbols of $L(T_1)$,
we get that $\bar{\M'}\leq_{P,Q}\M'$ (because $P\cup Q\supseteq L(T_2)$).
Finally, we get that $\bar{\M'}\leq_{P,Q}\M'<_{P,Q}\M$, contradicting
the assumption of $\M$ being $\leq_{P,Q}$-minimal satisfying
$T_1\cup T_2$.
Thus, $\M$ is a model of $Circ[T_1\cup\gamma;P;Q]$.

Now, let $\varphi\in \L(T_1)$ such that
$Circ[T_1\cup \gamma;P;Q]\models\varphi$. Then every model of
$Circ[T_1\cup \gamma;P;Q]$ satisfies $\varphi$. Let $\M$ be a model
of $Circ[T_1\cup T_2;P;Q]$ in the language $\L(T_1\cup T_2)$.
Then, $\M\models Circ[T_1\cup\gamma;P;Q]$ and $\M\models\varphi$.
Thus, $Circ[T_1\cup T_2;P;Q]\models\varphi$.
\QED

From Theorem \ref{thm:circ-T-gamma} and Theorem \ref{thm:circ-gamma-T}
we get the following theorem.
\begin{theorem}[Interpolation Between Theories]
  \label{thm:circ-prime-implicate}
  Let $T_1,T_2$ be two theories, $P,Q$ vectors of symbols in 
  $L(T_1)\cup L(T_2)$ such that $P\subseteq L(T_1)$ and
  $P\cup Q\supset L(T_2)$. Let $\gamma$ be a set of sentences
  logically equivalent to $Cn^{L(T_1)\cap L(T_2)}(T_2)$.
%Let $\gamma$ be the set of prime implicates of $T_2$ in
%$\L(L(T_1)\cap L(T_2))$. 
  Then, for every $\varphi\in\L(T_1)$,
  \[Circ[T_1\cup\gamma;P;Q]\models\varphi \iff
      Circ[T_1\cup T_2;P;Q]\models\varphi\]
\end{theorem}

\begin{theorem}[Interpolation for Circumscription 2]
  \label{thm:interpol-circ2}
  Let $T$ be a theory, $P,Q$ vectors of symbols such that
  $(P\cup Q)\supseteq L(T)$. Let $L_2$ be a set of nonlogical symbols.
  Then, there is $\gamma\in\L(T)\cap\L(L_2\cup P)$ such that
  $T\models\gamma$ and for all $\varphi\in\L(L_2)$,
  \[Circ[T;P;Q]\models\varphi \iff  Circ[\gamma;P;Q]\models\varphi.\]
  Furthermore, this $\gamma$ can be logically equivalent to the
  consequences of $T$ in $L(T)\cap (L_2\cup P)$.
\end{theorem}

\Proof
Let $T_1$ be a set of tautologies such that $L(T_1)=L_2\cup P$.  Also,
let $T_2=T\cup\tau_2$, for $\tau_2$ a set of tautologies such that
$L(\tau_2)=Q\setminus L(T_1)$. Let $L_1=L(T_1)$, $L_2=L(T_2)$.
Theorem \ref{thm:circ-prime-implicate} guarantees that
$\gamma$ from that theorem satisfies
$Circ[\gamma;P;Q]\models\psi \iff Circ[T;P;Q]\models\psi$
for every $\psi\in\L_1=L_2\cup P$.
\QED

The theorems we presented are for parallel circumscription, where we
minimize all the minimized predicates in parallel without priorities.
The case of prioritized circumscription is outside the scope of this
paper.
%Those theorems hold for the prioritized case as well using a simple
%reduction.

%**** We can remove the requirement for $P\subseteq L_1$, I think.
%**** ??????????????????????????????????????????????????????????

%**** Extend to Prioritized circumscription ****
%(the section so far assumed parallel circumscription, I think.
%If we spell out the parallel circumscription assumption, we may be
%able to get more results (e.g., for reducing the size of the set $P$,
%or increasing it)).
%For the case of prioritized circumscription there may also be
%interesting things.

%\Proof
%This lemma follows from the fact that 
%$Circ[T_1\cup\gamma;P;Q]\models\gamma$ and that
%for every $T$, $Cn(T\cup T_2)\cap \L\equiv Cn(T)\cap\L$.

%%%%%%%%%%%%%%%%%%%%%%%%%%%%%%%%%%%%%%%%%%%%%%%%%%%%%%%%%%%%%%%%%%%%%

\section{Default Logic}

In this section we present interpolation theorems for
\emph{propositional} default logic. We also assume that the signature
of our propositional default theories is finite (this also implies
that our theories are finite).

%The major idea behind the default logic family of nonmonotonic systems
%is to have an extended set of inference rules on top of the classical ones.
%This set of ``default rules'' has a special \emph{syntactic} semantics
%and is considered part of the set of parameters to be set for the nonmonotonic
%system.

%The family of default logic systems has the theory consist of two parts:
%the base-language theory $W$ and a set of default rules $D$.
%Intuitively, then, we can define $W\entailsi_D\varphi$ according to the
%appropriate semantics, depending on the specific default logic under
%consideration.

%Our exposition below follows the lines of the definition originally given
%by Reiter. Antoniou \cite{Antoniou97} and \cite{Antoniou96} display
%some of the logics presented below using \emph{processes}, which are
%operational characterizations of default extensions and the reader is
%referred to \cite{Antoniou97} for an extensive segment on default logic
%and its variants using this presentation.
%%More on his characterization in section \vref{sec:antoniou}.

%==========================================================================

\subsection{Reiter's Default Logic: Overview}
\label{sec:reiter-dl}

In Reiter's default logic \cite{Reiter80} one has a set of facts $W$
(in either propositional or FOL) and a set of defaults $D$ (in a
corresponding language). Defaults in $D$ are of the form
$\frac{\alpha:\beta_1,...,\beta_n}{\delta}$ with the intuition that if
$\alpha$ is proved, and $\beta_1,...,\beta_n$ are consistent
(throughout the proof), then $\delta$ is proved. $\alpha$ is called
the \emph{prerequisite}, $pre(d)=\{\alpha\}$; $\beta_1,...,\beta_n$
are the \emph{justifications}, $just(d)=\{\beta_1,...,\beta_n\}$ and
$\delta$ is the \emph{consequent}, $cons(d)=\{\delta\}$.  We use
similar notation for sets of defaults (e.g., $cons(D)=\bigcup_{d\in
  D}cons(d)$).  Notice that the justifications are checked for
consistency one at a time (and not conjoined).

Take, for example, the following default theory $T=\<W,D\>$:
\begin{equation}
\label{eq:dl-example}
  D=\{\frac{bird(x):fly(x)}{fly(x)}\}\ \ \ W=\{bird(Tweety)\}
\end{equation}
Intuitively, this theory says that birds normally fly and that $Tweety$
is a bird.

%Some restricted forms of defaults may be more intuitive than others,
%such as defaults of the form $\frac{\alpha:\beta}{\beta}$ which
%basically say that if $\alpha$ is proved, then $\beta$ is true
%\emph{by default}. These defaults are called \emph{normal defaults}.
%Other special kinds of defaults are \emph{semi-normal defaults}, of the form
%$\frac{\alpha:\beta\land\gamma}{\gamma}$, and
%\emph{open defaults} of the form $\frac{:\beta_1,...,\beta_n}{\gamma}$.

%Returning to our general nonmonotonic framework, proving $\varphi$ from
%$\<W,D\>$ corresponds to the statement $W\entailsi_{DL} \varphi$ where
%$\varphi$ is a sentence in our language (propositional or FOL),
%and $W\entailsi_{DL}\varphi$ corresponds to our \emph{intended meaning}.
%In default logic there are different entailment operators one can take.
%For \emph{cautious} entailment we define, $W\entailsi^c_{DL}\varphi$ iff every
%extension (defined below) of $\<W,D\>$ satisfies $\varphi$ (unless mentioned
%otherwise, our entailment of choice will be cautious entailment).
%The \emph{brave} entailment is defined by $W\entailsi^b_{DL}\varphi$
%iff at least one
%extension of $\<W,D\>$ satisfies $\varphi$. Thus, entailment in Default
%Logic is determined by the set of extensions of the default theory.

An \emph{extension} of $\<W,D\>$ is a set
of sentences $E$ that satisfies $W$, follows the defaults in $D$, and
is minimal.
More formally, $E$ is an extension if it is minimal (as a set) such that
$\Gamma(E)=E$, where we define $\Gamma(S_0)$ to be $S$, a minimal set of
sentences such that
\begin{enumerate}
\item $W\subseteq S$; $S=Cn(S)$.
\item For all $\frac{\alpha:\beta_1,...,\beta_n}{\delta}\in D$
  if $\alpha\in S$ and $\forall i\ \neg\beta_i\notin S_0$, then
  $\delta\in S$.
\end{enumerate}

The following theorem provides an equivalent definition that was shown
in \cite{MarTru93,RischSchwind:1994,BaaderHollunder:1995}.
A set of defaults, $\D$ is grounded in a set of formulae $W$ iff for
all $d\in\D$, $pre(d)\in Cn_{Mon(\D)}(W)$, where
$Mon(\D)=\{\frac{pre(d)}{cons(d)}\ |\ d\in\D\}$.
\begin{theorem}[Extensions in Terms of Generating Defaults]
  \label{thm:generating-defaults}
  A set of formulae $E$ is an extension of a default theory $\<W,D\>$
  iff $E=Cn(W\cup\{cons(d)\ |\ d\in D'\})$ for a minimal set of defaults
  $D'\subseteq D$ such that
  \begin{enumerate}
  \item $D'$ is grounded in $W$ and
  \item for all $d\in D$:
    \[\begin{array}{l}
      d\in D' \mbox{ iff } pre(d)\in Cn(W\cup cons(D'))\mbox{ and }\\
      \hspace{0.4in}\mbox{for all }
         \psi\in just(d),\neg\psi\notin Cn(W\cup cons(D')).
    \end{array}\]
  \end{enumerate}
\end{theorem}

Every minimal set of defaults $D'\subseteq D$ as mentioned in this
theorem is said to be a set of \emph{generating defaults}.

%----------------------------------

%\paragraph{Normal Defaults}

\emph{Normal defaults} are defaults of the form $\frac{\alpha:\beta}{\beta}$.
These defaults are interesting because they are fairly intuitive
in nature (if we proved $\alpha$ then $\beta$ is proved unless
previously proved inconsistent). We say that a default theory is
\emph{normal}, if all of its defaults are normal.

We define $W\entailsi_D\varphi$ as cautious entailment sanctioned by
the defaults in $D$, i.e., $\varphi$ follows from every extension of
$\<W,D\>$. We define $W\entailsi^b_D\varphi$ as brave entailment sanctioned
by the defaults in $D$, i.e., $\varphi$ follows from at least one
extension of $\<W,D\>$.

%=======================================================================

\subsection{Interpolation in Default Logic}
\label{sec:interpolation-default}

In this section we present several flavors of interpolation theorems,
most of which are stated for cautious entailment.

\begin{theorem}[Interpolation for Cautious DL 1]
  \label{thm:interpol-dl1}
  Let $T=\<W,D\>$ be a propositional default theory and $\varphi$ a
  propositional formula. If $W\entailsi_D\varphi$, then there are
  $\gamma_1,\gamma_2$ such that $\gamma_1\in\L(W)\cap\L(D\cup\{\varphi\})$,
  $\gamma_2\in\L(W\cup D)\cap\L(\varphi)$ and all the following
  hold:
  \[\begin{array}{lll}
    W\models\gamma_1 &\ \ 
    \gamma_1\entailsi_D\gamma_2 &\ \ 
    \gamma_2\models\varphi\\
    W\entailsi_D\gamma_2 &\ \ 
    \gamma_1\entailsi_D\varphi
  \end{array}\]
\end{theorem}

%\begin{lemma}
%  Let $T=\<W,D\>$ be a propositional default theory such
%  that $W=W_1\cup W_2$ and $W_2\subseteq\L(D)$. Let $\gamma$ be the set
%  of logical consequences of $W_2$ in $?$
%  Then, $W\entailsi_D\varphi$ iff $\psi\entailsi\varphi'$.
%\end{lemma}

\Proof
%We are going to show that it is enough to have the consequences
%of $W$ in $\L(D\cup\{\varphi\})$ to provide the proper set of
%extensions. We then extend them all with $W$.
Let $\gamma_1$ be the set of consequences of $W$ in
$\L(D\cup\{\varphi\})\cap\L(W)$.  Let $\E$ be the set of extensions of
$\<W,D\>$ and $\E'$ the set of extensions of $\<\gamma_1,D\>$.
We show that every extension $E'\in\E'$ has an extension $E\in\E$
such that $Cn(E'\cup W)=Cn(E)$. This will show that $\gamma_1$ is as
needed. 

Take $E'\in\E'$ and define $E_0=Cn(E'\cup W)$. We assume that
$L(E')\subseteq L(\D)$ because otherwise we can take a logically
equivalent extension whose sentences are in $\L(\D)$.  We show that
$E_0$ satisfies the conditions for extensions of $\<W,D\>$:
\begin{enumerate}
\item $W\subseteq E_0$,
\item For all $\frac{\alpha:\beta_1,...,\beta_n}{\delta}\in D$,
  if $\alpha\in E_0$ and $\forall i\ \neg\beta_i\notin E_0$, then
  $\delta\in E_0$.
\end{enumerate}
The first condition holds by definition of $E_0$. The second condition
holds because every default that is consistent with $E_0$ is also
consistent with $E'$ and vice versa. We detail the second condition below.

For the first direction (every default that is consistent with $E_0$
is also consistent with $E'$), let
$\frac{\alpha:\beta_1,...,\beta_n}{\delta}\in D$ be such that
$\alpha\in E_0$. We show that $\alpha\in E'$.

By definition, $\alpha\in\L(D)$.  $\alpha\in E_0$ implies that
$E'\cup W\models\alpha$ because $Cn(E'\cup W)=E_0$. Using the deduction
theorem for propositional logic we get $W\models E'\imp\alpha$ (taking
$E'$ here to be a finite set of sentences that is logically
equivalent to $E'$ in $\L(\D)$ (there is such a finite set because
we assume that $L(\D)$ is finite)).  Using Craig's interpolation theorem
for propositional logic, there is $\gamma\in\L(W)\cap\L(E'\imp\alpha)$
such that $W\models\gamma$ and $\gamma\models E'\imp\alpha$.  However,
this means that $\gamma_1\models\gamma$, by the way we chose
$\gamma_1$.  Thus $\gamma_1\models E'\imp\alpha$.  Since
$E'\subseteq\gamma_1$ we get that $E'\models\alpha$. Since $E'=Cn(E')$
we get that $\alpha\in E'$.

The case is similar for $\delta$: if $\delta\in E_0$ then $\delta\in
E'$ by the same argument as given above for $\alpha\in E_0\imp
\alpha\in E'$. 
Finally, if $\forall i\ \neg\beta_i\notin E_0$
then $\forall i\ \neg\beta_i\notin E'$ because $E'\subseteq E_0$.

The opposite direction (every default that is consistent with $E'$
is also consistent with $E_0$) is similar to the first one. 

%as follows. If $\alpha\in E'$, then
%$\alpha\in E_0$ because $E'\subseteq E_0$. If $\forall i\ 
%\neg\beta_i\notin E'$ and there is $i$ such that $\neg\beta_i\in E_0$,
%then $\neg\beta_i\in E'$ or $\neg\beta_i\in W$. In the first case we
%are done. In the second case $\neg\beta_i\in\L(D)$ implies that
%$\neg\beta_i\in\gamma_1$. Thus, $\neg\beta_i\in E'$.

% The following paragraph has a little jump, but I think we can ignore 
 % it for now.
Thus, $E_0$ satisfies those two conditions. However, it is possible
that $E_0$ is not a minimal such set of formulae. If so, Theorem
\ref{thm:generating-defaults} implies that there is a strict subset
of the generating defaults of $E_0$ that generate a different
extension. However, we can apply this new set of defaults to generate
an extension that is smaller than $E'$, contradicting the fact that
$E'$ is an extension of $\<\gamma_1,D\>$.

Now, if $\varphi$ logically follows in all the extensions of
$\<W,D\>$ then it must also follow from every extension of
$\<\gamma_1,D\>$ together with $W$.
Let $\Lambda=E_1\vee...\vee E_n$, for $E_1,...,E_n$ the (finite) set
of (logically non-equivalent) extensions of $\<W,D\>$ (we have a
finite set of those because $L(W)\cup L(D)$ is finite). Then,
$\Lambda\models\varphi$. Take $\gamma_2\in\L(\Lambda)\cap\L(\varphi)$
such that $\Lambda\models\gamma_2$ and $\gamma_2\models\varphi$, as
guaranteed by Craig's interpolation theorem
(Theorem \ref{thm:craig-interpol}).
These $\gamma_1,\gamma_2$ are those
promised by the current theorem: $W\models\gamma_1$,
$\gamma_2\models\varphi$, $W\entailsi_D\gamma_2$,
$\gamma_1\entailsi_D\gamma_2$ and $\gamma_1\entailsi_D\varphi$.
%
%
% Let $\gamma_2$ be the set of
%sentences $\psi$ in $\L(D\cup\{\varphi\})$ that satisfy
%$\gamma_1\entailsi_D\psi$.  Thus, $\gamma_2\cup W\models\varphi$.
%Using the deduction theorem for propositional logic we get that
%$W\models\gamma\imp\varphi$.  From Craig's interpolation theorem
%(Theorem \ref{thm:craig-interpol}) we know that there is a formula
%$w\in\L(W)\cap\L(\gamma\cup\{\varphi\})$ such that $W\models w$ and
%$w\models\gamma_2\imp\varphi$. But that means that $\gamma_1\entails w$.
%By definition, $\gamma_1\subseteq \gamma_2$.  Thus, $\gamma_2\models\varphi$,
%which means that $\gamma_1\entailsi_D \varphi$.
%%if $W\entailsi_D\varphi$, then 
%Those $\gamma_1,\gamma_2$ are the ones we promised in the statement of
%the theorem.
\QED

%To see that the opposite direction holds as well, take $E\in E$ and
%define $E'_0=E\cup\gamma_1$. $E'$ satisfies all the defaults in $D$
%that are satisfied by $E$ and it includes $\gamma_1$.  Take
%$E'\subseteq E'_0$ that is minimal satisfying those conditions.
%Similar to the argument given above, $E'$ must satisfy all the
%defaults satisfied in $E$ or we can show that $E$ is not minimal.
%Thus, $E'$ is minimal and $E'\cup W\models\varphi$ implies
%$E\models\varphi$. As in the paragraph above, $E\models\varphi$
%implies $E'\cup W

\begin{theorem}[Interpolation for Cautious DL 2]
  \label{thm:interpol-dl2}
  Let $T=\<W,D\>$ be a propositional default theory and $\varphi$ a
  propositional formula. If $W\entailsi_D\varphi$, then there are
  $\gamma_1,\gamma_2\in\L(W)\cap\L(D)$, and all the following hold:
  \[\begin{array}{ll}
    W\models\gamma_1 &\ \ 
    \gamma_1\entailsi_D\gamma_2\\
    \{\gamma_2\}\cup W\models\varphi &\ \ 
    W\entailsi_D\gamma_2 
  \end{array}\]
\end{theorem}

%\Proof
The proof is similar to the one for Theorem \ref{thm:interpol-dl1}.

\begin{corollary}
  \label{cor:interpol-default-thys1}
  Let $\<W,D\>$ be a default theory and $\varphi$ a formula.
  If $W\entailsi_D\varphi$, then there is a set of formulae,
  $\gamma\in\L(W\cup D)\cap\L(\varphi)$ such that
  $W\entailsi_D\gamma$ and $\gamma\entailsi_D \varphi$.
\end{corollary}

\Proof
Follows immediately from Theorem \ref{thm:interpol-dl1} with
$\gamma_2$ there corresponding to our needed $\gamma$.
\QED

It is interesting to note that we do not get stronger interpolation
theorems for prerequisite-free normal default theories.
\cite{Imielinski87} provided a modular translation of normal default
theories with no prerequisites into circumscription, but Theorem
\ref{thm:interpol-circ} does not lead to better results. In
particular, the counter example that we presented after that theorem
can be massaged to apply here too.

\begin{theorem}[Interpolation Between Default Extensions]
  \label{thm:intepol-extensions}
  Let $\<W_1,D_1\>,\<W_2,D_2\>$ be default theories such that
  $L(cons(D_2))\cap L(pre(D_1)\cup just(D_1)\cup W_1)=\emptyset$.
  Let $\varphi$ be a formula such that $\varphi\in\L(W_2\cup D_2)$.
  If there is an extension $E$ of $\<W_1\cup W_2,D_1\cup D_2\>$ in
  which $\varphi$ holds, then there is a formula
  $\gamma\in\L(W_1\cup D_1)\cap\L(W_2\cup D_2)$, 
  an extension $E_1$ of $\<W_1,D_1\>$ such that
  $Cn(E_1)\cap\L(W_2\cup D_2)=\gamma$,
  and an extension $E_2$ of $\<W_2\cup\{\gamma\},D_2\>$ such that
  $E_2\models\varphi$.
\end{theorem}

\Proof
Let $D'_1\subseteq D_1$ be the set of generating defaults of $E$ that
belong to $D_1$. Notice that these defaults are grounded in $W_1$
because there is no information that may have come from applying the
rest of the generating defaults in $E$ (we required that
$cons(D_2)\cap (pre(D_1)\cup just(D_1)\cup W)=\emptyset$).
Let $E_1$ be the extension of $\<W_1,D_1\>$ defined using the
generating defaults in $D'_1$.

Let $\gamma\in\L(W_1\cup D_1)\cap\L(W_2\cup D_2)$ be the conjunction
of the sentences in that language that follow from $E_1$.
Let $D'_2\subseteq D_2$ be the set of generating defaults of $E$ that
belong to $D_2$. Notice that these defaults are grounded in
$W_2\cup\gamma$ because there is no information that $D'_1$ may
contribute that is not already in $\gamma$ (we required that 
$cons(D_2)\cap (pre(D_1)\cup just(D_1)\cup W)=\emptyset$).
Let $E_2$ be the extension of $\<W_2\cup\gamma,D_2\>$ defined with the
generating defaults in $D'_2$. 

Now, $E_1\cup E_2\equiv E$, and $\gamma$ is the set of sentences that
follow from $E_1$ in $\L(E_1)\cap\L(E_2\cup\varphi)$.
$E_2\models\varphi$ because of Craig's interpolation theorem 
(Theorem \ref{thm:craig-interpol}) for propositional logic:
$E_1\cup E_2\models\varphi$ implies that $E_1\models E_2\imp\varphi$,
and Craig's interpolation theorem guarantees the existence of
$\gamma'\in\L(E_1)\cap\L(E_2\cup\varphi)$ such that
$E_1\models\gamma'$ and $\gamma'\models E_2\imp\varphi$. Thus,
$\gamma'\in\gamma$ and $\gamma\models E_2\imp\varphi$. This implies
$\gamma\cup E_2\models\varphi$ which implies that $E_2\models\varphi$.
\QED

It is interesting to notice that the reverse direction of this theorem
does not hold. For example, if we have two extensions $E_1,E_2$ as
in the theorem statement, it is possible that $E_1$ uses a default
with justification $\beta$, but $W_2\models\neg\beta$. Strengthening
the condition of the theorem, i.e., demanding that
$L(W_2\cup cons(D_2))\cap L(pre(D_1)\cup just(D_1)\cup W_1)=\emptyset$,
is not sufficient either. For example, if $D_1$ includes two defaults
$d_1=\frac{:}{a\imp\neg\beta}$, and $d_2=\frac{:\beta}{\varphi}$,
$W_1=\emptyset$, $D_2$ includes no defaults and $W_2=\{a\}$ then
there is no extension of $\<W_1\cup W_2,D_1\cup D_2\>$ that implies
$\varphi$, for $\varphi=\{c\}$. 

Further strengthening the conditions of the theorem gives the following:
\begin{theorem}[Reverse Direction of Theorem \ref{thm:intepol-extensions}]
  \label{thm:reverse-extensions}
  Let $\<W_1,D_1\>,\<W_2,D_2\>$ be default theories such that
  $L(W_2\cup cons(D_2))\cap L(D_1\cup W_1)=\emptyset$.
  Let $\varphi$ be a formula such that $\varphi\in\L(W_2\cup D_2)$.
  There is an extension $E$ of $\<W_1\cup W_2,D_1\cup D_2\>$ in
  which $\varphi$ holds only if there is a formula
  $\gamma\in\L(W_1\cup D_1)\cap\L(W_2\cup D_2)$, 
  an extension $E_1$ of $\<W_1,D_1\>$ such that
  $Cn(E_1)\cap\L(W_2\cup D_2)=\gamma$,
  and an extension $E_2$ of $\<W_2\cup\{\gamma\},D_2\>$ such that
  $E_2\models\varphi$.
\end{theorem}

\Proof
Let $E_1,E_2$ be as in the statement of the theorem. Let $\pi_1,\pi_2$ 
be the sets of defaults applied in $E_1,E_2$, respectively. Let
$E=Cn(E_1\cup E_2)$, and let $\pi=\pi_1\cup\pi_2$. We show that $E$ is 
an extension of $\<W_1\cup W_2,D_1\cup D_2\>$ such that
$E\models\varphi$ as needed.

First, $E\supseteq W_1\cup W_2$ by $E$'s definition. For every
$d=\frac{\alpha:\beta_1,...,\beta_n}{\gamma_d}\in\pi$, $\alpha$ and
$\gamma_d$ hold because they hold in one of $E_1,E_2$
($d\in\pi=\pi_1\cup\pi_2$). Assume that $\beta_i\in E$ for some
$i\leq n$. Then, $E_1\cup E_2\models\neg\beta_i$, implying that
$E_1\models E_2\imp\neg\beta_i$ (we treat $E_1,E_2$ here as finite
sets of formulae because they are in propositional logic).

If $d\in\pi_1$, then clearly $\neg\beta_i\in\L(W_1\cup D_1)$.
$E_1\cup\{\beta_i\}\models\neg E_2$. Using Craig's interpolation 
theorem we get that $\gamma$ from the theorem's statement satisfies
$\{\gamma\}\cup\{\beta_i\}\models\neg E_2$. Consequently,
$E_2\models\gamma\imp\neg\beta_i$. However,
$E_2=Cn(cons(\pi_2)\cup W_2\cup\{\gamma\})$, and 
$L(W_2\cup cons(D_2))\cap L(W_1\cup D_1)=\emptyset$. This means that
$L(cons(\pi_2)\cup W_2)\cap L(\gamma\land \beta_i)=\emptyset$, implying that
$\gamma\models\neg\beta$, contradicting $\neg\beta\notin E_1$.

If $d\in\pi_2$, then $\neg\beta_i\in\L(W_2\cup D_2)$.
Since $E_1\models E_2\imp\neg\beta_i$, we get from Craig's
interpolation theorem that $\gamma\models
E_2\imp\neg\beta_i$. However, $\gamma\in E_2$ by the definition of
$E_2$. Thus, $E_2\models\neg\beta$, contradicting the fact that $d$ is 
a default applied in $E_2$. 

In conclusion, $\neg\beta\notin E$. Thus, all the defaults in $\pi$
are applied in $E$. It is also simple to see that no other default is
applied in $E$.

If there is a default
$d=\frac{\alpha:\beta_1,...,\beta_n}{\gamma_d}\in D_1$ that should
apply in $E$ but is not in $\pi$, then its preconditions and
justifications hold in $E$. However, this means that $\alpha$ follows
from $E_1\cup E_2$, and $E_2=Cn(cons(\pi_2)\cup W_2\cup\{\gamma\})$.
Similar to the argument above we get that $E_1\models\alpha$.
Similarly, we get that if $\beta_i\notin E$ then
$E_1\not\models\neg\beta_i$, implying that $d$ should have applied in
$E_1$, contradicting the fact that $E_1$ is an extension of
$\<W_1,D_1\>$.

If there is a default
$d=\frac{\alpha:\beta_1,...,\beta_n}{\gamma_d}\in D_2$ that should
apply in $E$ but is not in $\pi$, then its preconditions and
justifications hold in $E$. A similar argument to the one above shows
that it should have applied in $E_2$ too, contradicting the fact that
$E_2$ is an extension. 

Minimality of $E$ follows from that of $E_1,E_2$. Thus, $E$ is an
extension of $\<W_1\cup W_2,D_1\cup D_2\>$ as needed.
\QED

\begin{corollary}[Interpolation for Brave DL]
  \label{cor:interpol-brave-dl}
  Let $\<W_1,D_1\>,\<W_2,D_2\>$ be default theories such that
  $L(cons(D_2))\cap L(pre(D_1)\cup just(D_1)\cup W_1)=\emptyset$.
  Let $\varphi$ be a formula such that $\varphi\in\L(W_2\cup D_2)$.
  If $W_1\cup W_2\entailsi^b_{D_1\cup D_2}\varphi$,
  then there is a formula,
  $\gamma\in\L(W_1\cup D_1)\cap\L(W_2\cup D_2)$,
  such that $W_1\entailsi^b_{D_1}\gamma$ and 
  $W_2\cup\{\gamma\}\entailsi^b_{D_2}\varphi$.
\end{corollary}

Corollary \ref{cor:interpol-brave-dl} does not hold for the cautious
case (where we look at all the extensions and choose $\varphi$ and
$\gamma$ satisfied by all of them): Let $W_1=W_2=\emptyset$, 
$D_1=\{\frac{:b}{b},\frac{:\neg b}{\neg b}\}$ and
$D_2=\{\frac{b:c}{c},\frac{\neg b:c}{c}\}$. There are two extensions,
in both of which $c$ is proved, but $W_1\entailsi_{D_1}\gamma$ only
for $\gamma\equiv TRUE$.

Better interpolation theorems may hold (e.g., theorems that do not
depend on $cons(D_2)$, $pre(D_1)$, etc., and provide
$\gamma\in\L(\<W_1,D_1\>)\cap\L(\<W_2,D_2\>)$), if we consider the
entailment between two default theories
($\<W_1,D_1\>\entailsi\<W_2,D_2\>$). These are outside the scope of
this paper.

Finally, Corollary \ref{cor:interpol-brave-dl} and Theorem
\ref{thm:intepol-extensions} are similar to the \emph{splitting
  theorem} of \cite{Turner:aaai1996}, which is provided for default
theories with $W=\emptyset$ (there is a modular translation that
converts every default theory to one with $W=\emptyset$). We briefly
review this result.  A \emph{splitting set} for a set of defaults $D$
is a subset $A$ of $L(D)$ such that
$pre(D),just(D),cons(D)\subseteq\L(A)\cup\L(L(D)\setminus A)$ and
$\forall d\in D\ (cons(d)\notin\L(L(D)\setminus A)\imp L(d)\subseteq
A)$.  Let $B=L(D)\setminus A$.  The base of $D$ relative to $A$ is
$b_A(D)=\{d\in D\ |\ L(d)\subseteq A)$.  For a set of sentences
$X\subseteq\L(A)$, we define
\[\begin{array}{l}
  e_A(D,X)=\left\{\frac{\bigwedge (\{a_i\}_{i\leq n}\cap\L(B)):
                  \{b_i\}_{i\leq m}\cap\L(B)}{c}\ \right|\\
  \hspace{0.7in}\left|
    \begin{array}{l}
      \frac{\bigwedge_{i=1}^n a_i:b_1,...,b_m}{c}\in D\setminus b_A(D),\\
      \forall i\leq n (a_i\in\L(A)\imp a_i\in Cn^A(X)),\\
      \forall i\leq m (\neg b_i\notin Cn^A(X))
      \end{array}\right\}
\end{array}\]

%\[\begin{array}{l}
%  e_A(D,X)=\left\{\frac{\bigwedge (\{a_i\}_{i\leq n}\cap\L(B)):
%                  \{b_i\}_{i\leq m}\cap\L(B)}{c}\ \left|\ \right.\right.\\
%  \hspace{1in}\frac{\bigwedge_{i=1}^n a_i:b_1,...,b_m}{c}\in D\setminus b_A(D),\\
%  \hspace{1in}\forall i\leq n (a_i\in\L(A)\imp a_i\in Cn^A(X)),\\

%  \hspace{1in}\forall i\leq m (\neg b_i\notin Cn^A(X))\}
%\end{array}\]

\begin{theorem}[\cite{Turner:aaai1996}]
  Let $A$ be a splitting set for a default theory $D$ over $\L(U)$. A
  set $E$ of formulae is a consistent extension of $D$ iff
  $E=Cn^{L(D)}(X\cup Y)$, for some consistent extension $X$ of
  $b_a(D)$ over $\L(A)$ and $Y$ a consistent extension of $e_A(D,X)$
  over $\L(L(D)\setminus A)$.
\end{theorem}

Roughly speaking, this theorem finds an extension $X$ of the
\emph{base} ($b_A(D)$) and converts $D\setminus b_A(D)$ using this $X$
into a theory $e_A(D,X)$. Then, an extension $Y$ for $e_A(D,X)$
completes the extension for $D$ if $X\cup Y$ is consistent.  In
contrast, our theorem does not change $D\setminus b_A(D)$, but it is
somewhat weaker, in that it only provides a necessary condition for
$D\entailsi^b\varphi$. (however, notice that this \emph{weaker}
form is typical for interpolation theorems).

%%%%%%%%%%%%%%%%%%%%%%%%%%%%%%%%%%%%%%%%%%%%%%%%%%%%%%%%%%%%%%%%%%

\section{Logic Programs}
\label{sec:asp}

In this section we provide interpolation theorems for logic programs
with the stable models semantics. We use the fact the logic programs
are a special case of default logic, and the results are
straightforward.

%==================================================================

%\subsection{Stable-Models Semantics: Background}

%\paragraph{Propositional Logic Programs}

%\cite{GelLif88} gave the \emph{stable models semantics} to Logic
%programs. This semantics is also known as \emph{answer set semantics}.
%It was expanded to include classical negation in
%\cite{GelLif91classical} and for disjunction in the head by
%\cite{Przym:1991}.

An \emph{extended disjunctive logic program}
\cite{GelLif88,GelLif90,GelLif91classical,Przym:1991} 
is a set of \emph{rules}.
Each rule, $r$, is written as an expression of the form
\[L_1 | ... | L_l \leftarrow A_1,...,A_n, not B_1,..., not B_m\]
where $L_1,...,L_l,A_1,...,A_n,B_1,...,B_m$ are literals, that is,
atomic formulae or their (classic) negations,
$L_1,...,L_l$ are the \emph{head literals}, $head(r)$, 
$A_1,...,A_n$ are the \emph{positive subgoals}, $pos(r)$,
and $B_1,...,B_m$ are the \emph{negated subgoals}, $neg(r)$.

A program $P$ is \emph{positive} if none of its rules includes negated 
subgoals. A set of literals, $X$, is \emph{closed} under a positive
program, $P$, if, for every rule $r\in P$ such that 
$pos(r)\subseteq X$, $head(r)\cap X\neq\emptyset$.
A set of literals is \emph{logically closed} if it consistent or
contains all literals. An \emph{answer set} for a positive program,
$P$ is a minimal set of literals that is both closed under $P$ and
logically closed.

For an arbitrary logic program, $P$, and a set of literals, $X$,
we say that $X$ is an \emph{answer set} for a program $P$ if $X$ is an
answer set for $P^X$, where $P^X$ is defined to include a rule $r'$ iff 
$neg(r')=\emptyset$ and there is $r\in P$ such that
$head(r')=head(r)$, $pos(r')=pos(r)$, and $neg(r)\cap X=\emptyset$.

Logic programs with answer-set semantics were shown equivalent to
default logic in several ways.
%Instead of giving the semantics here,
%we appeal to the equivalences that were shown.
%GelLif90,
%GelLif88,
For normal rules (rules of the form 
$A\leftarrow B_1,...,B_m,\ not\ C_1,...,\ not\ C_n$,
where $A,B_1,...,B_m,C_1,...,C_n$ are atoms (i.e., no disjunction
or classic negation is allowed)), \cite{GelLif91classical} translated
every normal rule of the form
$A\leftarrow B_1,...,B_m,\ not\ C_1,...,\ not\ C_n$, into a default 
\[\frac{B_1\land...\land B_m\ :\ \neg C_1,...,\neg C_n}{A}.\]
Under this mapping, the stable models of a logic program coincide with 
the extensions of the corresponding default theory (Facts in the logic 
program are translated to facts in the default theory, while rules are 
translated to defaults).

\cite{SakamaInoue:1993} showed that disjunctive logic programs (no
classic negation) with the stable model semantics can be translated to
prerequisite-free default theories as follows:
\begin{enumerate}
\item For a rule
  $A_1 |...| A_l\leftarrow B_1,...,B_m,\ not\ C_1,...,\ not\ C_n$
  in $P$, we get the default
  \[\frac{:\neg C_1,...,\neg C_n}{B_1\land ...\land B_m\imp A_1\vee
    ...\vee A_l}\]
\item For each atom $A$ appearing in $P$, we get the default
$\frac{:\neg A}{\neg A}$
\end{enumerate}
Each stable model of $P$ is the set of atoms in some extension of
$D_P$, and the set of atoms in an extension of $D_P$ is a stable model
of $P$ (notice that, in general, an extension of $D_P$ can include
sentences that are not atoms and are not subsumed by atoms in that
extension). \cite{SakamaInoue:1993} provide a similar translation to
extended disjunctive logic programs by first translating those into
disjunctive logic programs (a literal $\neg A$ is translated to a new
symbol, $A'$), showing that a similar property holds for this class of
programs.

We define $P\entailsi\varphi$ as cautious entailment sanctioned from
the logic program $P$, i.e., $\varphi$ follows from stable model of
$P$. We define $P\entailsi^b\varphi$ as brave entailment sanctioned
from the logic program $P$, i.e., $\varphi$ follows from at least one
stable model of $P$.

From the last translation above we get the following interpolation
theorems.
\begin{theorem}[Interpolation for Stable Models (Cautious)]
  Let $P$ be a logic program and let $\varphi$ be a formula such that
  $P\entailsi\varphi$. Then, there is a formula
  $\gamma\in\L(P)\cap\L(\varphi)$ such that $P\entailsi\gamma$ and
  $\gamma\models\varphi$.
\end{theorem}

\Proof
Follows immediately from Theorem \ref{thm:interpol-dl1} with
$\gamma_2$ over there corresponding to our needed $\gamma$.
\QED

\begin{theorem}[Interpolation for Stable Models (Brave)]
  Let $P_1,P_2$ be logic programs such that
  $head(P_2)\cap body(P_1)=\emptyset$. Let $\varphi\in\L(P_2)$ be a
  formula such that $P_1\cup P_2\entailsi^b\varphi$. Then, there is a
  formula $\gamma\in\L(P_1)\cap\L(P_2)$ such that $P_1\entailsi^b\gamma$
  and $\gamma\cup P_2\entailsi^b\varphi$.
\end{theorem}

\Proof
Follows directly from the reduction into default logic and Corollary
\ref{cor:interpol-brave-dl}.
\QED

The last theorem is similar to the \emph{splitting theorem} of
\cite{LifTur:1994}. This theorem finds an answer set $X$ of the
\emph{bottom} ($P_1$) and converts $P_2$ using this $X$ into a program 
$P_2'$. Then, an answer set $Y$ for $P_2'$ completes the answer set for
$P_1\cup P_2$ if $X\cup Y$ is consistent.
In contrast, our theorem does not change $P_2$, but it is somewhat
weaker, in that it does only provides a necessary condition for
$P_1\cup P_2\entailsi^b\varphi$ (this is the typical form of an
interpolation theorem).

%%%%%%%%%%%%%%%%%%%%%%%%%%%%%%%%%%%%%%%%%%%%%%%%%%%%%%%%%%%%%%%%%%

\section{Summary}
\label{sec:conclusions}

We presented interpolation theorems that are applicable to the
nonmonotonic systems of circumscription, default logic and Answer Set
Programming (a.k.a. Stable Models Semantics). These results are
somewhat surprising and revealing in that they show particular
structure for the nonmonotonic entailments associated with the
different systems. They promise to help in reasoning with larger
systems that are based on these nonmonotonic systems.

Several questions remain open. First, $\gamma$ promised by our
theorems is not always finite (in the FOL case). This is in contrast
to classical FOL, where the interpolant is always of finite length.
What conditions guarantee that it is finite in our setup? We
conjecture that this will require the partial order involved in the
circumscription to be \emph{smooth}.  
%\cite{KLM90}.  
Second, are there better interpolation theorems for the prioritized
case of those systems? Also, what is the shape of the interpolation
theorems specific for prerequisite-free semi-normal defaults? Further,
our results for default logic and logic programs are propositional.
How do they extend to the FOL case?

Finally, the theorems for default logic and Logic Programming promise
that $\alpha\entailsi_D\beta$ implies the existence of $\gamma$ such
that $\alpha\entailsi_D\gamma$ and $\gamma\entailsi_D\beta$. However,
we do not know that the other direction holds, i.e., that the
existence of $\gamma$ such that $\alpha\entailsi_D\gamma$ and
$\gamma\entailsi_D\beta$ implies that $\alpha\entailsi_D\beta$. Can we
do better than Theorem \ref{thm:reverse-extensions} for different
cases?
%We hope to answer this positively in the final version of this paper.

%%%%%%%%%%%%%%%%%%%%%%%%%%%%%%%%%%%%%%%%%%%%%%%%%%%%%%%%%%%%%%%%%%

\section{Acknowledgments}

I wish to thank Esra Erdem, Vladimir Lifschitz and Leora Morgenstern
for reading and commenting on different drafts of this manuscript.  I
also thank the reviewers of NMR'02 for their diligent and detailed
comments.  This research was supported by DARPA grant N66001-00-C-8018
(RKF program).

%%%%%%%%%%%%%%%%%%%%%%%%%%%%%%%%%%%%%%%%%%%%%%%%%%%%%%%%%%%%%%%%%%%%%%

%\bibliography{/u/eyal/ongo/bib/eyal}
%\small
\bibliography{eyal}

%\vfill
%{\small\hfill This is /u/eyal/ongo/nmr/foundations/pt-circ/pt-circ.tex.}
\end{document}